\documentclass[sigconf]{acmart}

\usepackage{booktabs} % For formal tables
\usepackage{algorithm}
\usepackage{algorithmic} %format of the algorithm

\usepackage{subfigure}
\usepackage{graphics}
\usepackage{ifthen}
\usepackage{latexsym}
\usepackage{CJK}
\usepackage{amsfonts}
\usepackage{ifthen}
\usepackage{caption}
\usepackage{graphicx}
\usepackage{float}
\usepackage{multirow}
\usepackage{subfigure}
\usepackage{balance}
\usepackage{subfigure}
\usepackage{graphics}
\usepackage{enumerate}
\usepackage{verbatim}
\usepackage{graphicx}
\usepackage{float}
\usepackage{booktabs}
\usepackage[hang,flushmargin]{footmisc}
\usepackage{caption}
\usepackage{bbold}
\usepackage{color}
\usepackage{xspace}

\usepackage{listings}
\usepackage{xcolor}
\usepackage{multicol}
\usepackage{adjustbox}
\usepackage{calc}

\usepackage{booktabs}    % 专业表格线
\usepackage{makecell}    % 多行单元格支持
\usepackage{multirow}    % 多行合并
\usepackage{graphicx}    % 缩放支持(备用)

\definecolor{codegreen}{rgb}{0,0.6,0}
\definecolor{codegray}{rgb}{0.5,0.5,0.5}
\definecolor{codepurple}{rgb}{0.58,0,0.82}

\newcommand{\paratitle}[1]{\vspace{1.5ex}\noindent\textbf{#1}}

\newcommand{\ignore}[1]{}

\AtBeginDocument{%
  }

\copyrightyear{2025}
\acmYear{2025}
\setcopyright{acmlicensed}\acmConference[WWW Companion '25]{Companion Proceedings of the ACM Web Conference 2025}{April 28-May 2, 2025}{Sydney, NSW, Australia}
\acmBooktitle{Companion Proceedings of the ACM Web Conference 2025 (WWW Companion '25), April 28-May 2, 2025, Sydney, NSW, Australia}
\acmDOI{10.1145/3701716.3715512}
\acmISBN{979-8-4007-1331-6/2025/04}

\newcommand{\baby}{\textsc{EBaReT}\xspace}
\begin{document}

\title{EBaReT: Expert-guided Bag Reward Transformer for Auto Bidding}

\author{Kaiyuan Li}
\authornote{Both authors contributed equally to this research.}
\affiliation{%
  \institution{Kuaishou Technology}
  \city{Beijing}
  \country{China}}
\email{likaiyuan03@kuaishou.com}

\author{Pengyu Wang}
\affiliation{%
  \institution{Kuaishou Technology}
  \city{Beijing}
  \country{China}}
\authornotemark[1]
\email{wangpengyu03@kuaishou.com}

\author{Yunshan Peng}
\affiliation{%
  \institution{Kuaishou Technology}
  \city{Beijing}
  \country{China}}
\email{pengyunshan@kuaishou.com}

\author{Pengjia Yuan}
\affiliation{%
  \institution{Kuaishou Technology}
  \city{Beijing}
  \country{China}}
\email{yuanpengjia@kuaishou.com}

\author{Yanxiang Zeng}
\affiliation{%
  \institution{Kuaishou Technology}
  \city{Beijing}
  \country{China}}
\email{zengyanxiang@kuaishou.com}
 
\author{Rui Xiang}
\affiliation{%
  \institution{Kuaishou Technology}
  \city{Beijing}
  \country{China}}
\email{xiangrui@kuaishou.com}

\author{Yanhua Cheng}
\affiliation{%
  \institution{Kuaishou Technology}
  \city{Beijing}
  \country{China}}
\email{chengyanhua@kuaishou.com}

\author{Xialong Liu}
\affiliation{%
  \institution{Kuaishou Technology}
  \city{Beijing}
  \country{China}}
\email{zhaolei16@kuaishou.com}

\author{Peng Jiang}
\affiliation{%
  \institution{Kuaishou Technology}
  \city{Beijing}
  \country{China}}
\email{jiangpeng@kuaishou.com}

\renewcommand{\shortauthors}{Kaiyuan Li et al.}

\begin{abstract}
Reinforcement learning has been widely applied in automated bidding. Traditional approaches model bidding as a Markov Decision Process (MDP). Recently, some studies have explored using generative reinforcement learning methods to address long-term dependency issues in bidding environments. Although effective, these methods typically rely on supervised learning approaches, which are vulnerable to low data quality due to the amount of sub-optimal bids and low probability rewards resulting from the low click and conversion rates. Unfortunately, few studies have addressed these challenges. 

In this paper, we formalize the automated bidding as a sequence decision-making problem and propose a novel \textbf{E}xpert-guided \textbf{Ba}g \textbf{Re}ward \textbf{T}ransformer (\baby) to address concerns related to data quality and uncertainty rewards. Specifically, to tackle data quality issues, we generate a set of expert trajectories to serve as supplementary data in the training process and employ a Positive-Unlabeled (PU) learning-based discriminator to identify expert transitions. To ensure the decision also meets the expert level, we further design a novel expert-guided inference strategy. Moreover, to mitigate the uncertainty of rewards, we consider the transitions within a certain period as a "bag" and carefully design a reward function that leads to a smoother acquisition of rewards. Extensive experiments demonstrate that our model achieves superior performance compared to state-of-the-art bidding methods.
\end{abstract}

\begin{CCSXML}
<ccs2012>
   <concept>
       <concept_id>10002951.10003227.10003447</concept_id>
       <concept_desc>Information systems~Computational advertising</concept_desc>
       <concept_significance>500</concept_significance>
       </concept>
 </ccs2012>
\end{CCSXML}

\ccsdesc[500]{Information systems~Computational advertising}

\keywords{Decision Transformer, Auto Bidding, Reinforcement Learning}
\maketitle
\section{Introduction}
\begin{figure}[ht]
  \centering
  \includegraphics[scale=0.23]{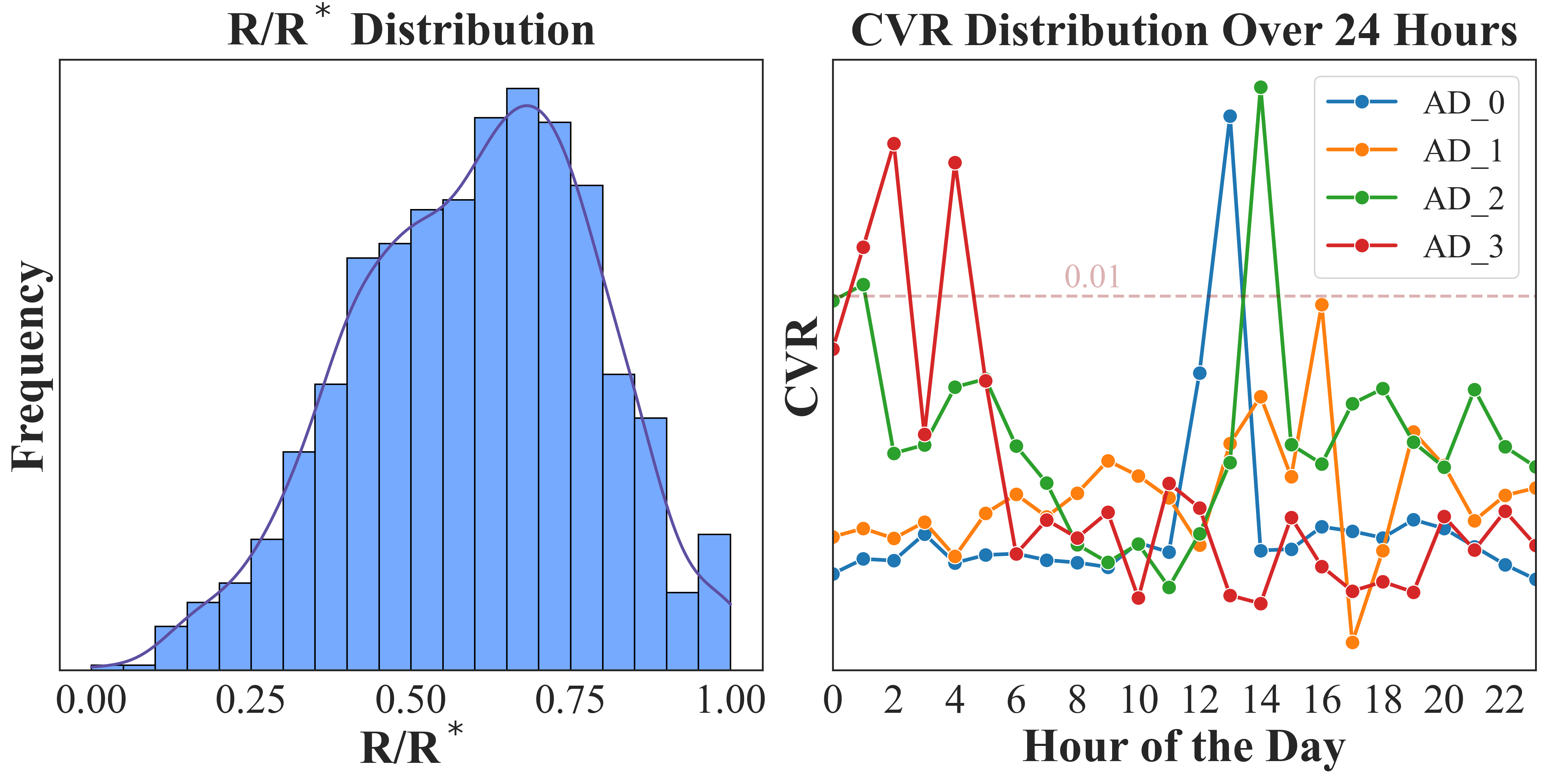}
  \caption{The figure on the left illustrates the ratio distribution between advertisers' online conversions $R$ and their optimal offline conversions $R^*$. In contrast, the figure on the right depicts the variation in four top advertisers' conversion rates throughout the day.}
  \label{fig:intro}
\end{figure}
Auto-bidding has become an essential interface between advertisers and internet advertising, gaining significant traction by allowing advertisers to articulate high-level goals and constraints\cite{auto_bidding_suvery}. Recently, reinforcement learning has been widely applied in online advertising bidding\cite{auto_bid_1,auto_bid_2}, framing the bidding problem as a competition-based reinforcement learning to develop an effective bidding policy. However, most approaches model the bidding problem as a Markov Decision Process (MDP)\cite{rl_bid_2,rl_bid_3,rl_bid_4,rl_bid_5}, and ignore the impact of long-term dependencies in the bidding environment. To address this, Guo et al.\cite{real_aigb} introduced a generative approach to model automated bidding as a sequential decision-making process, achieving promising results.

However, the direct application of generative algorithms in real-world advertising bidding scenarios presents two significant challenges: \textbf{1) Low Quality of logged Data}. As illustrated on the left side of Figure \ref{fig:intro}, we observe many suboptimal decisions, with actual conversions $R$ significantly lower than offline optimal conversions $R^*$. Applying supervised methods is inevitably hindered by noisy behaviors. \textbf{2) Low Probability of Reward Acquisition}. As shown on the right side of Figure \ref{fig:intro}, even for top advertisers, the conversion rate remains very low (i.e., <0.01) which means that even if the advertiser places an optimal bid and wins the auction, they may still fail to receive effective reward feedback in the uncertain environment which complicates the optimization process.

To tackle these challenges, inspired by Decision Transformer (DT)\cite{dt} which formalizes RL as conditional sequence modeling, we propose a novel approach, \textbf{E}xpert-guided \textbf{Ba}g \textbf{Re}ward \textbf{T}ransformer (\baby) to address the above concerns. To address data quality issues, we generate a set of expert trajectories for each advertiser based on the truthful optimal theory. To identify expert behavior, rather than treating all offline data as negative examples, we employ a Positive-Unlabeled (PU) learning approach to develop a discriminator that distinguishes between expert and non-expert behavioral strategies. We integrate these trajectories using the learned discriminator to differentiate them at various expert levels during training and condition decision-making on the highest expert behavior to ensure decisions meet expert levels during testing. To address the environmental uncertainty problem, we define bidding behaviors over a specific period as a "Bag" based on our statistical analysis and design a novel reward redistribution strategy using the learned discriminator to enhance the short-term credit assignments. Our contributions can be summarized as follows.
\begin{itemize}
\item We propose a novel framework for generative bidding and focus on the two key issues of generative models in real bidding scenarios: low data quality and low probability rewards.
\item To address data quality issues, we introduce an expert-guided training and inference method. We generate a set of expert trajectories and learn a discriminator to distinguish the expert transitions during training, relying solely on expert knowledge for inference during testing.
\item To alleviate environmental uncertainty, we propose a Bag-based reward redistribution strategy based on the carefully designed reward function and then dynamically modify the Return-to-Go (RTG) during the training process.
\item We validate the effectiveness of our approach through extensive experiments, with both offline and online results confirming the effectiveness of our approach.

\end{itemize} 

\section{Preliminary}
\paratitle{Bidding and Auction}.
 Consider the environment with $n$ auto bidding agents, indexed by $i$, and $m$ auctions, indexed by $j$. The valuation of agent $i$ winning auction $j$ is $v_{ij} \in R^+$. value $v_{ij}$ may include the click-through-rate~(CTR) and/or the conversion-rate~(CVR). In each auction $j$, each agent submits a bid $b_{ij} \in R^+$ and the auction takes the vector of bids $b_{j}=(b_{1j},b_{2j}, \cdots, b_{nj})$ as the input and determines the allocations $x_{ij}(b_{ij})\in[0,1]$ and payments $p_{ij}(b_{j}) \in R$ for each agent $i$.

\paratitle{Bidding Problem}.
In the application of online advertising, the bidding problem is usually to maximize
a given objective while subject to some constraints. The constraints most commonly studied are the budget and the \textit{return-on-spend (RoS)} constraints. We can formulate the bidding agent's problem using the following linear programming (LP) function:
\begin{align*}
\max & \sum_{j} x_{ij} \cdot v_{ij}  \tag{Bidding}\\
\text { s.t. } & \sum_{j} (p_{ij} \cdot b_i) \leq B_i  \tag{Budget}\\
& \frac{\sum_{j} (x_{ij} \cdot p_{ij})}{\sum_{j} (x_{ij} \cdot v_{ij})} \leq C_{i} \tag{RoS}
\end{align*}
\section{Methodology}
In this section, we present the proposed \baby in detail, with the overall architecture illustrated in Figure 1.
\begin{figure*}[htbp]
  \centering
  \includegraphics[scale=0.34]{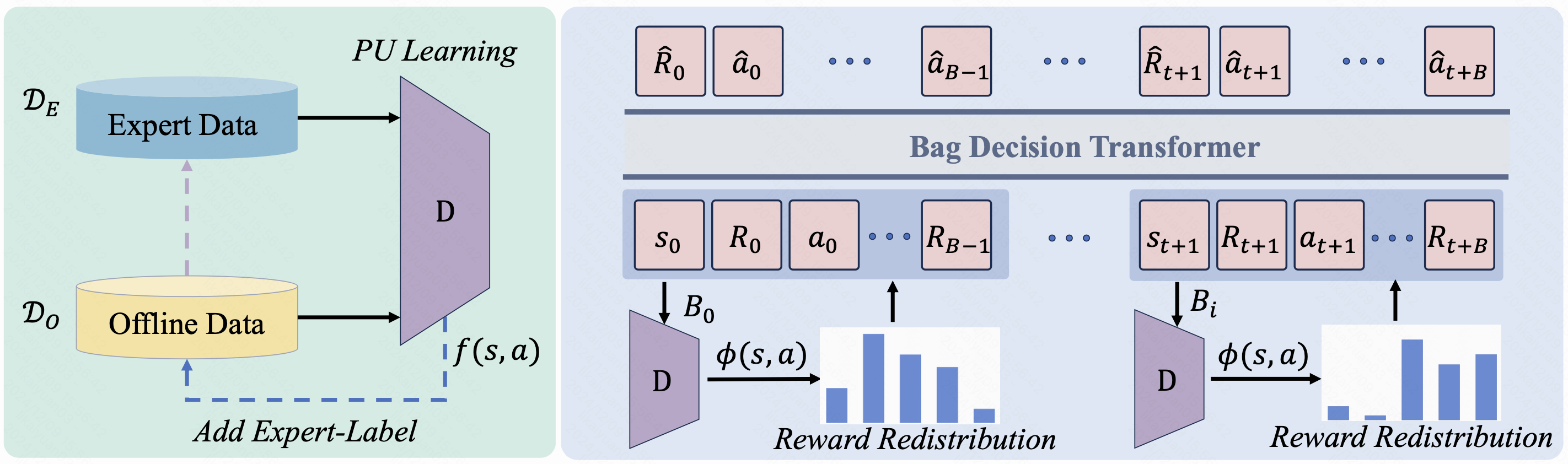}
  \caption{The overall architecture of \baby. We first generate expert trajectories and train a discriminator, after this, we train a bag transformer by reward-distribution strategy. Finally, we make decisions with an expert-guided strategy.}
\end{figure*}
\subsection{Bag Decision Transformer}
\paratitle{Task Definition}: We consider a decision-making agent for each advertiser that, at each time step $t$, receives a state $s_t$ of the bidding environment (e.g., remaining budget, predicted CTR/CVR, etc.) and selects an action $a_t$ as bid and obtains an immediate reward $r_t$ such as the number of conversions. Our objective is to learn an optimal bidding strategy $a^*_t = f_{\theta} \left(s_{\leq t}, a_{<t}, R_{<t}\right)$ with parameters $\theta$, which maximizes the agent’s RTG $R_t = \sum_{k>t} r_k$ in bidding environments.

\paratitle{Trajectory Representation}: 
As illustrated in Figure \ref{fig:intro}, the conversion rate of advertisers shows a significant temporal trend. Hence, we partitioned each day into multiple consecutive time intervals, called a "Bag". Following the work of \cite{multi-game, fr}, the bags and sequences we consider take the following form:
\begin{equation}
B_{i} = \left\langle s_t, R_t, a_t \ldots, s_{t+B-1}, R_{t+B-1}, a_{t+B-1} \right\rangle, \tau =\left\langle\ \ldots, B_i, \ldots\right\rangle
\end{equation}
% \begin{equation}
% \tau =\left\langle\ \ldots, B_j, \ldots\right\rangle
% \end{equation}
Unlike DT, this framework enables the prediction of the RTG without the manual selection of an expert-level return during inference. For each token in the trajectory, we introduced a bag position embedding for each element to enhance the representation of tokens within the same bag.

\paratitle{Loss Function}: We employ two distinct autoregressive loss functions to supervise two individual tasks. First, we sample $(s, a)$ from the data $\mathcal{D}$ and predict the current time step's RTG using \( \hat{R}_t = f_{\theta}(\ldots, s_t) \), which resembles the Q-function. Next, we predict the action for the next time step using \( \hat{a}_t = f_{\theta}(\ldots, s_t, R_t) \).
\begin{equation}\label{eq:r_loss}
\mathcal{L}\left(\theta\right) =  \sum_{(s,a) \in \mathcal{D}}\left( \hat{R}_t-  R_t\right)^2 + \sum_{(s,a) \in \mathcal{D}} \left( \hat{a}_t-  a_t\right)^2
\end{equation}

\subsection{Expert Generation and Discriminator}
We contend that both expert and non-expert trajectories contribute to improved learning outcomes. Specifically, leveraging large and diverse datasets that encompass both expert and non-expert trajectories enhances performance.
\subsubsection{Expert Trajectory Generation}
To generate expert trajectories, we omit the subscript of $i$ as we will take the perspective of a single bidding agent while assuming all other agents are fixed. We rewrite the LP function to its Dual LP. The bidding
formula results in an auction outcome identical to an optimal primal solution $b_j$, if the underlying auction is truthful \cite{auto_bidding_suvery}. For each query $j$:
\begin{equation}
b_j = \frac{1 + \alpha_{c} \cdot C}{\alpha_{b} + \alpha_{c}} \cdot v_j 
\end{equation}
where $\alpha_{b}$ represents the influence of budget constraints and $\alpha_{c}$ denotes the impact of RoS constraints.

\subsubsection{Expert Transition Discriminator}
We construct a discriminator $d(s, a)$ to discern between expert and sub-optimal transitions. To avoid designating all offline transitions as negative, we opt for PU learning during the discriminator training process. As a result, we consider the offline dataset as an unlabeled set with the expert dataset serving as the labeled positive dataset. We use a non-negative risk estimator described in\cite{nnPU} to make the discriminator more robust:
\begin{equation}
\begin{aligned}
\min_d \quad & \eta \underset{(s, a) \sim \mathcal{D}_E}{\mathbb{E}} \left[-\log \sigma(d(s, a))\right] \\
& + \max \left( 0, \underset{(s, a) \sim \mathcal{D}_O}{\mathbb{E}} \left[-\log \left(1 - \sigma(d(s, a))\right)\right] \right. \\
& \left. - \eta \underset{(s, a) \sim \mathcal{D}_E}{\mathbb{E}} \left[-\log \left(1 - \sigma(d(s, a))\right)\right] \right)
\end{aligned}
\end{equation}
Here, $\eta$ is a hyperparameter that represents the positive class prior. $\mathcal{D}_E$ is the expert dataset and $\mathcal{D}_O$ is the offline logged dataset. 
\subsection{Bag-based Reward Redistribution}
The \baby is trained to ensure that, for each reward bag, the sum of predicted rewards aligns with the total bagged reward. Given the trained discriminator, Inspired by Inverse RL\cite{irl_suvery}, we design a reward relocated function
\begin{equation}\label{eq:phi_r}
\phi(r|s_t,a_t) = \exp \frac{d(s_t,a_t)}{\beta}
\end{equation}
where parameter $\beta$ controls the shape of the reward distribution. According to the equation \ref{eq:phi_r}, we can calculate the reward for each state-action tuple in a bag as follows:
\begin{equation}
\hat{r}_t = \frac{\phi(r|s_t,a_t)}{\sum_{j \in B_i} \phi(r|s_j,a_j)} \left( \sum_{r_j\in B_i } r_j \right)
\end{equation}
At each time step, transitioning from time $t$ to $t+1$, we adjust the input RTG as follows:
\begin{equation}
R_{t+1} = R_{t} - \hat{r}_t
\end{equation}
Instead of subtracting the actual reward, we redistribute the rewards within the bag, leading to a smoother acquisition of rewards rather than relying solely on sporadic conversions.
\subsection{Expert-guided Action Inference}
As previously described, our training datasets encompass both expert and non-expert behaviors. However, directly generating action based on the imitation of mixed data is unlikely to obtain expert-level behavior according to the equation \ref{eq:best_q}.
\begin{equation}\label{eq:best_q}
\mathbb{E}_{s_0 \in \mathcal{D}_{O}}{[R_0]}  < \mathbb{E}_{s_0 \in \mathcal{D}_{E}}{[R_0]}
\end{equation}
Inspired by \cite{multi-game,rcp}, we discrete the distance function $d(s, a)$ into $k$ levels and introduce an expert token to represent different levels of expert transition. We can then utilize this token to differentiate the sources of the transitions during training and always assign the expert token as the highest expert level during testing. Through it, we transform a conditional reasoning problem in a continuous space into a k-ary conditional reasoning process and ensure the decision meets the expert level.
\section{Experiment}
\paratitle{Experimental Environment}.
The simulated experimental environment is conducted in a manually built offline real advertising system\cite{auction_net} released by Alibaba\footnote{https://github.com/alimama-tech/AuctionNet}. The dataset includes 21 advertising periods from 7 to 27, each containing more than 500,000 impression opportunities and divided into 48 steps. Each opportunity includes 50 agents with the bids. The dataset comprises over 500 million records. Each record includes the predicted conversion value, bid, auction, and impression results. For a fair comparison, we use the data from periods 7-13 as the training dataset and the data from periods 14-20 as the test dataset. 

\paratitle{Experimental Setting}.
We compare \baby with two state-of-the-art traditional offline RL methods containing IQL\cite{iql}, CQL\cite{cql}, and three generative methods containing BC\cite{bc}, DT\cite{dt}, and DiffBid\cite{kdd_2024}. For evaluation, we use cumulative conversions as the evaluation metric. For our model, one bag contains 8 time steps. $\eta$ in PU learning is set to 0.01, $\beta$ in the reward function is set to 0.5, and the number of expert levels is set to 2.

\subsection{Performance Evaluation}
In this section, we compare the performance of \baby with various baselines. The overall performance is reported in Table~\ref{t:overall}.
\begin{table}[!t]
  \caption{Performance comparison between baselines and \baby . The best performance of each column is underlined.}
  \label{t:overall}
  \centering
  \begin{tabular}{llllllll}
    \toprule
    % \cmidrule(l){1-2}
    Method     & P14     & P15 & P16 & P17 & P18 & P19 & P20 \\
    \midrule
    CQL     & 31.09 & 29.99  & 29.88  & 29.73  & 27.97  & 33.45  & 27.99     \\
    IQL     & 32.75 & 25.43  & 27.05  & 35.09  & 30.38  & 34.07  & 28.39     \\
    BC & 28.53  & 27.37  & 30.55  & 28.64  & 29.13 & 31.70  & 26.43   \\
    DT     & 30.01 & 29.79  & 30.31  & 31.92  & 29.52  & 34.42  & 29.70     \\
    DiffBid     & 31.95 & 28.62  & 30.65  & 35.59  & 27.72  & 34.36  &  29.79    \\
    \baby     & \underline{37.89} & \underline{33.31} & \underline{36.09} & \underline{40.40} & \underline{33.42} & \underline{40.66} & \underline{33.71} \\
    \bottomrule
  \end{tabular}
\end{table}
We found that among traditional RL methods, IQL performed the best, which is similar to the findings in \cite{real_aigb}. For generative methods, we discovered that DT outperformed BC. Our model achieved the best results, validating our proposed approach's effectiveness.
\subsection{Ablation Study}
We tested our proposed expert-guided training and testing method and compared it with several variant methods: 1) \baby$_{\neg{E}}$: The optimizations related to expert data were removed, and the model degenerated to DT. 2) \baby$_{\neg{PU}}$: All offline data was treated as negative examples, causing the loss to degenerate into cross-entropy. 3) \baby$_{\neg{EA}}$: We remove the expert-guided inference method and remove the expert token.
4) \baby$_{\neg{BR}}$: We remove the reward redistribution strategy. The experiments confirmed the effectiveness of each design module.
\begin{table}[htbp]
  \caption{Comparisons between \baby and its four variant models. The best performance is underlined.}
  \label{tb:infer}
  \centering
  \begin{tabular}{l@{\hskip 0.2cm}l@{\hskip 0.2cm}l@{\hskip 0.2cm}l@{\hskip 0.2cm}l@{\hskip 0.2cm}l@{\hskip 0.2cm}l@{\hskip 0.2cm}l}
    \toprule
    Method & P14 & P15 & P16 & P17 & P18 & P19 & P20 \\
    \midrule
    \baby & \underline{37.89} & 33.31 & \underline{36.09} & \underline{40.40} & \underline{33.42} & \underline{40.66} & 33.71 \\
    \baby$_{\neg{E}}$ & 30.44 & 30.53 & 32.62 & 31.75 & 31.18 & 34.66 & 28.80 \\
    \baby$_{\neg{PU}}$ & 37.26 & 33.71 & 33.25 & 36.12 & 32.99 & 38.37 & 31.14 \\
    \baby$_{\neg{EA}}$ & 34.92 & 31.86 & 30.80 & 34.32 & 30.78 & 33.97 & 29.79 \\
    \baby$_{\neg{BR}}$ & 36.17 & \underline{35.41} & 34.89 & 35.91 & 33.41 & 38.98 & \underline{34.23} \\
    \bottomrule
  \end{tabular}
\end{table}
\subsection{Online Test}
\baby is designed for real-world auto bidding. We also want to check whether it can benefit the online environment. We conducted an online test for online service from 2024-11-19 to 2024-11-25. The baseline is similar to IQL and the experiment result shown in Table~\ref{t:online} demonstrates the effectiveness of our approach.
\begin{table}[!t]
\caption{Performance of \baby in Online Testing.}
\label{t:online}
\centering % Center the table
\begin{tabular}{l|cc|cc}
\toprule
Method & Budget& Plan & Conversions & Revenue \\ 
\midrule
Baseline                & 138,369 &113                       & 565,795         & 5,753,090   \\ 
Ours                   & 138,369 &113                    & 641,970          & 6,411,794    \\ 
\textit{compare}       & - &-                      & +13.46\%         & +11.44\% \\ 
% Ours without expert                   & 138369                     & 18          & 155    \\ 
% Ours without                    & 138369                     & 18          & 155    \\ 

\bottomrule
\end{tabular}
\end{table}

\section{Conclusion}
In this paper, we formalize the auto-bidding problem as a sequential decision-making task and propose \baby to address the issues of low data quality and low probability reward acquisition. Specifically, we construct expert trajectories and design a novel PU Learning-based discriminator to identify expert transitions. We then design an expert-guided decision-making strategy during testing to ensure the action is expert-level. To mitigate environmental uncertainty, we further design a Bag-based reward redistribution mechanism. For future work, we consider introducing a more unified training and inference framework that does not rely on manual priors to implement long-term and short-term reward redistribution strategies.

\bibliographystyle{ACM-Reference-Format}
\bibliography{sample-base}

\end{document}